# Learning-based Control for Tendon-Driven Continuum Robotic Arms

Nima Maghooli *, Omid Mahdizadeh, Mohammad Bajelani and S. Ali A. Moosavian

Center of Excellence in Robotics and Control, Advanced Robotics and Automated Systems (ARAS), Department of Mechanical Engineering, K. N. Toosi University of Technology, Tehran, Iran.

## Abstract

Tendon-Driven Continuum Robots (TDCRs) are widely recognized for their flexibility and adaptability in constrained environments, making them invaluable for most applications, such as medical surgery, industrial tasks, and so on. However, the inherent uncertainties and highly nonlinear dynamics of these manipulators pose significant challenges for classical model-based controllers. Addressing these challenges necessitates the development of advanced control strategies capable of adapting to diverse operational scenarios. This paper presents a centralized position control of TDCRs using Deep Reinforcement Learning (DRL), with a particular focus on the Sim-to-Real transfer of control policies. The proposed method employs a customized Modified Transpose Jacobian (MTJ) control strategy for continuum arms, where its parameters are optimally tuned using the Deep Deterministic Policy Gradient (DDPG) algorithm. By integrating an optimal adaptive gain-tuning regulation, the research aims to develop a model-free controller that achieves superior performance compared to ideal model-based strategies. Both simulations and real-world experiments demonstrate that the proposed controller significantly enhances the trajectory-tracking performance of continuum manipulators. Applied on a variety of initial conditions and paths, effectively establishes a promising controller for any task.

## 1 Introduction

The advancement of TDCRs presents a significant opportunity to enhance precision and adaptability in applications such as medical devices, flexible manufacturing systems, and exploratory robotics (Burgner-Kahrs et al., 2015). These robots, characterized by their continuous and flexible structures, offer superior dexterity compared to traditional rigid robots, making them ideal for navigating complex and constrained environments. However, their control remains a significant challenge due to their high degrees of freedom (DOF) and nonlinear dynamics (Chikhaoui and Burgner-Kahrs, 2018). Addressing these challenges requires innovative control strategies capable of handling the inherent complexities of TDCRs (George Thuruthel et al., 2018; Wang et al., 2021b).

Traditional control methods often struggle with the high-dimensional and nonlinear nature of TDCRs, necessitating the exploration of advanced techniques such as reinforcement learning (RL). RL has emerged as a promising approach due to its capacity to handle complex, high-DOF systems without requiring precise analytical models. For example, the application of DDPG in controlling a spatial three-section continuum robot demonstrated superior trajectory-tracking performance, with a maximum error of only 1 mm compared to traditional methods (Djeffal et al., 2024). Similarly, Fuzzy Reinforcement Learning (FRL) has shown effectiveness in trajectory tracking under varying conditions, leveraging robust Cosserat rod-based modeling (Goharimanesh et al., 2020). These studies underscore the potential of RL in overcoming the limitations of conventional controllers, particularly in tasks demanding high precision.

Beyond TDCRs, RL has been successfully applied in soft robotics to address similar challenges. Visual learning-based controllers, for instance, have been utilized for soft robotic fish, enabling flexible and cost-effective designs by reducing reliance on complex curvature-sensing electronics (Rajendran and Zhang, 2022). Similarly, model-free RL methods have been shown to enhance task performance in soft continuum robots across tasks like reaching, crank rotation, and peg-in-hole operations (Morimoto et al., 2022). These advancements highlight the adaptability and scalability of RL methods for various robotic systems.

One of the critical challenges in deploying RL-based controllers for continuum robots is the transition from simulation to real-world applications, known as Sim-to-Real transfer. This process is essential for validating RL algorithms in dynamic environments. Robust Sim-to-Real strategies have been developed to address this challenge. For instance, ELFNet demonstrated that policies trained in simulation could be directly applied to real-world scenarios with minimal performance degradation (Morimoto et al., 2021). Similarly, integrated tracking control approaches combining rolling optimization with RL have proven effective in real-time tasks, such as managing space debris (Jiang et al., 2022). These efforts emphasize the importance of incorporating model dynamics and reward design to enhance the stability and efficiency of RL algorithms in real-world applications.

Despite the progress, challenges such as noise processes, state representation, and training stability persist. Comparative studies have shed light on these aspects, providing insights for fine-tuning RL frameworks to improve their robustness and adaptability (Kołota and Kargin, 2023). By addressing these challenges, RL-based controllers can bridge the gap between theoretical advancements and practical utility, paving the way for broader deployment in tendon-driven continuum robots and other high-DOF robotic systems.

Based on previous studies, three general strategies have been considered for employing Deep Reinforcement Learning (DRL) in control systems (Wang et al., 2021a). The first strategy involves utilizing the DRL agent as a model-free intelligent controller and force distributor, where the agent directly learns the control policy and parameters without relying on any analytical models of the system's kinetics or kinematics. Although this approach allows for the full utilization of DRL's capacity, it requires extensive training time and computational resources. The second strategy also employs the DRL agent as a model-free intelligent controller; however, it incorporates the system's kinematics (e.g., Jacobian matrix) for force distribution. While this strategy reduces complexity compared to the first one, it still imposes a significant computational burden. In contrast, the third strategy, adopted in this study, leverages the DRL agent as an adaptive gain-tuning system for a model-free controller. Instead of directly acting as the primary controller, the DRL agent optimizes the controller's parameters under various operating conditions, effectively combining the strengths of analytical models (e.g., kinematics) with DRL's optimization capabilities.

Previous works in this domain have primarily focused on the first and second strategies, where DRL-based controllers were designed to fully or partially replace analytical models. However, the approach in this study follows the third strategy, which takes advantage of both analytical modeling and DRL to achieve efficient and adaptive parameter tuning. Given that extracting forward kinematics equations and calculating the Jacobian matrix for a TDCR is straightforward and can be achieved with reasonable accuracy, completely bypassing this information in favor of a fully model-free DRL approach is unnecessary. Furthermore, utilizing DRL as the main controller is computationally expensive and time-intensive, making it less suitable for real-time applications. Therefore, the selected strategy prioritizes the integration of available analytical models with DRL to achieve a more efficient and practical control solution.

The main contribution of this paper is the proposed Sim-to-Real transfer of control policies, enabling the MTJ controller to achieve precise trajectory tracking starting from any arbitrary initial condition and following any desired trajectory in the workspace. Consequently, in this research, DRL is employed as an optimal adaptive gain-tuner system, providing the following advantages:

- Reduced Training Time through Predefined Structure: The time required for training the DRL agent is reduced because the agent benefits from a pre-fixed control structure. Therefore, the algorithm's effort is only focused on finding the optimal parameters.

- Locally Robust Control with Adaptive Gains: The proposed approach leverages DRL as a gain-tuning mechanism for locally robust PID-based controller (MTJ). This ensures that the control strategy remains robust in local operating conditions while adapting the controller gains optimally for dynamic system requirements.

- Lightweight Neural Network for Real-Time Application: The algorithm results in a simplified and efficient neural network policy, which is lightweight enough to facilitate real-time implementation on hardware systems without compromising performance.

- Improved Stability during Learning: Prior knowledge on the systems and the controller's parameters helps us to reduce the training time. The primary goal of designing a controller with adaptable and suitable gains for tracking any desired trajectory in the task-space of the system is achieved.

The paper is organized as follows: The Introduction highlights the motivation, background, and relevance of learning-based control for TDCRs. The Basic Discussions section details the simulation model, Jacobian Matrix Null-Space Analysis, and the DRL Application in Control System Design. The Proposed Learning-based Controller section integrates the MTJ control strategy with the DDPG algorithm, explaining the strategy and implementation. The Obtained Results section demonstrates the learning process and simulation outcomes, while the Experimental Implementation section validates the approach through real-world tests. The Discussion analyses results, comparing them with similar methods (e.g., Fuzzy-MTJ), and explores potential improvements. Finally, the Conclusions summarize key findings and suggest future research directions. Table 1 provides descriptions of the symbols used in the article.

**TABLE 1**
Nomenclature.

| Symbol | Definition | Symbol | Definition |
|---|---|---|---|
| $e_i$ | Position error of the end-effector along the $i$-axis in the task-space | $\dot{\boldsymbol{L}}$ | Vector of tendon length change rates in the continuum robotic arm |
| $\boldsymbol{e}$ | Position error vector of the end-effector in the task-space | $\boldsymbol{p}$ | Position of a point on the arc |
| $\dot{e}_i$ | Velocity error of the end-effector along the $i$-axis in the task-space | $s$ | Backbone reference length parameter |
| $\dot{\boldsymbol{e}}$ | Velocity error vector of the end-effector in the task-space | $T$ | Transpose of the vector or matrix |
| $e_{max_i}$ | Sensitivity threshold of the position error of the end-effector along the $i$-axis in the task-space for the MTJ controller | $T_i$ | Tension force of tendon $i$ |
| $\dot{e}_{max_i}$ | Sensitivity threshold of the velocity error of the end-effector along the $i$-axis in the task-space for the MTJ controller | $\boldsymbol{T}$ | Vector of generalized forces in joint space (tendon tensions) |
| $\boldsymbol{h}$ | Vector of system dynamics estimator terms in the MTJ controller | $\boldsymbol{T}^+$ | Vector of generalized forces in joint space (tendon tensions) after passing the null-space projection operator |
| $\boldsymbol{I}$ | Identity matrix | $t$ | Time |
| $\boldsymbol{J}$ | Linear Jacobian matrix (direct mapping of linear velocities from joint space to task-space) | $\boldsymbol{X}$ | Task-space variables vector (end-effector position vector) |
| $\boldsymbol{K}$ | Coefficient matrix of the system dynamics estimator in the MTJ controller | $\dot{\boldsymbol{X}}$ | Task-space velocities vector (end-effector velocity vector) |
| $\boldsymbol{K}_D$ | Derivative coefficient matrix of the MTJ controller | $\boldsymbol{\eta}$ | Non-trivial solution to the linear algebraic system |
| $\boldsymbol{K}_I$ | Integral coefficient matrix of the MTJ controller | $\boldsymbol{\zeta}$ | Null-space adjustment vector ($\boldsymbol{\zeta} \in \mathbb{R}^6$) |
| $\boldsymbol{K}_P$ | Proportional coefficient matrix of the MTJ controller | $\boldsymbol{\mathcal{F}}$ | Vector of generalized forces in the task-space |
| $l_i$ | Tendon length of tendon $i$ in the continuum robotic arm | $\dagger$ | Pseudo-inverse of a non-square matrix |
| $\boldsymbol{L}$ | Vector of tendon lengths in the continuum robotic arm | | |

## 2 Basic Discussions

In this section, the kinematics and dynamics models of the robot are presented, which are used to derive the Jacobian matrix and simulate the system's behavior in the simulation environment. Subsequently, the null space of the Jacobian matrix is analyzed, forming the basis for enhancing the MTJ algorithm for controlling continuum robots. Finally, the application of DRL in control system design is examined using the DDPG algorithm.

### 2.1 Kinematics and Kinetics Modeling

Given the primary focus of this study on the position control of the end-effector within the task-space, forward kinematics refers to the direct mapping from the joint-space to the task-space of the TDCR (passing through the configuration-space). The kinetics model for continuum robots can be categorized as either a dynamics model or a statics model. The dynamics model is a memory-based model and is used for accelerating movements. In contrast, the statics model is memory-less, making it suitable for quasi-static movements (Grassmann et al., 2022; Rucker and Webster, 2011).

This study employs a statics model with the assumption of constant curvature for each subsegment, known as the Piecewise Constant Curvature (PCC) model, selected for its effectiveness in representing continuum robots (Yuan et al., 2019; Rao et al., 2021). The PCC model was used for training the DRL agent and simulating the proposed controller. Compared to the Variable Curvature (VC) model, the PCC model requires significantly less computational cost while maintaining adequate accuracy. The VC model considers the dependence on time ($t$) and backbone reference length parameter ($s$) for the instantaneous position of each point along the backbone ($\boldsymbol{p} = f(t, s)$), offering high precision. However, this approach results in a set of nonlinear Partial Differential Equations (PDEs), which are computationally intensive and impractical for real-time applications (Dehghani and Moosavian, 2013). The PCC model simplifies these dependencies using two main assumptions (quasi-static motion and constant curvature for each subsegment), resulting in a set of nonlinear algebraic equations that are computationally less demanding and sufficiently accurate for the intended purposes.

### 2.2 TDCR Redundancy Resolution

One of the most critical issues in tendon-driven and cable-driven robotic systems is preventing tendon slack, or more precisely, maintaining tension in the tendons. This research utilizes the Null-Space Projection Operator (NSPO) of the Jacobian matrix to address this problem. The Jacobian matrix is a crucial tool for analyzing the structural characteristics of robotic systems. By calculating this matrix, the structural properties of tendon-driven continuum robots can be examined. Various methods have been proposed for computing the Jacobian, and in this section, the linear part of this matrix is derived from the forward kinematics equations (Jones and Walker, 2006). The continuum robot studied in this research is a two-segment system, with each segment actuated by three tendons. If the position vector of the end-effector is defined as $\boldsymbol{X} = [x \quad y \quad z]^T$ and the tendon length vector as $\boldsymbol{L} = [l_1 \quad l_2 \quad l_3 \quad l_4 \quad l_5 \quad l_6]^T$, the linear Jacobian matrix, which maps the rate of change between these two vectors ($\dot{\boldsymbol{X}} = \boldsymbol{J}\dot{\boldsymbol{L}}$), is computed as follows:

$$\boldsymbol{J}_{nm} = \frac{\partial X_n}{\partial L_m} \quad n \in \{1, \dots, \dim(\boldsymbol{X})\} \;\; , \;\; m \in \{1, \dots, \dim(\boldsymbol{L})\} \tag{1}$$

For the continuum robot under consideration, this results in a 3×6 rectangular matrix. The partial derivatives corresponding to the elements of this matrix are analytically derived using the forward kinematics equations. The null-space of matrix $\boldsymbol{J}$ is the set of all non-trivial solutions ($\boldsymbol{\eta}$) to the linear algebraic system $\boldsymbol{J\eta} = \boldsymbol{0}$. In the problem of position control for the end-effector of a tendon-driven continuum robot in the task-space, the number of system inputs (tendon tensions) exceeds the number of system outputs (end-effector position coordinates), making the system over-actuated with six inputs and three outputs.

In TDCRs, The mapping of generalized forces from the joint-space ($\boldsymbol{T} = [T_1 \quad T_2 \quad T_3 \quad T_4 \quad T_5 \quad T_6]^T$) to the task-space ($\boldsymbol{\mathcal{F}} = [F_x \quad F_y \quad F_z]^T$) is expressed as $\boldsymbol{\mathcal{F}} = \boldsymbol{J}^{-T}\boldsymbol{T}$ where $\boldsymbol{J}^{-T} = (\boldsymbol{J}^T)^{\dagger}$. Due to the non-square nature of $\boldsymbol{J}^T$, the left pseudo-inverse of this matrix is used. Using the projection operator in the null-space (Chiaverini et al., 2008; Godage et al., 2012), denoted as $[\boldsymbol{I} - \boldsymbol{J}^T\boldsymbol{J}^{-T}]$, the set of all solutions can be written as:

$$\boldsymbol{T} = \boldsymbol{J}^T\boldsymbol{\mathcal{F}} + [\boldsymbol{I} - \boldsymbol{J}^T\boldsymbol{J}^{-T}]\,\boldsymbol{\zeta} \tag{2}$$

In the above equation, $\boldsymbol{\zeta} \in \mathbb{R}^6$ is the null-space adjustment vector, and $[\boldsymbol{I} - \boldsymbol{J}^T\boldsymbol{J}^{-T}] \neq \boldsymbol{0}$. All vectors of the form $\boldsymbol{T}_n = [\boldsymbol{I} - \boldsymbol{J}^T\boldsymbol{J}^{-T}]\,\boldsymbol{\zeta}$ lie in the null-space of $\boldsymbol{J}^{-T}$. In other words, $\boldsymbol{T}_n \neq \boldsymbol{0}$, but the corresponding task-space force $\boldsymbol{\mathcal{F}}_n = \boldsymbol{J}^{-T}\boldsymbol{T}_n = \boldsymbol{0}$ (self –force).

### 2.3 DRL Application in Control System Design

The goal of solving a problem using DRL is to find an optimal mapping from the state-space to the action-space, known as the policy. The policy dictates the action to be taken by the DRL agent in each state, and the optimal policy aligns perfectly with the rewards received from the environment (Kober et al., 2013; François-Lavet et al., 2018). Initial efforts in development of DRL algorithms assumed a continuous state-space and a discrete action-space, such as the Deep Q-Network (DQN). The basis of this method is to assign a value to the state-action value function ($Q_{(s,a)}$) for each action in each state, and ultimately, a greedy action is selected for each state (Wu et al., 2020). The extension of this approach for continuous state and action-spaces is achieved by the DDPG algorithm. This algorithm utilizes deep neural networks to approximate the state-action value function and the policy. These network structures enable the DRL model to effectively map states to actions and evaluate the resultant action values, facilitating optimal policy learning. In temporal difference-based algorithms, the return is usually estimated by the value function. The state-action value function is defined as the expected return when in state $s_t$ and taking action $a_t$ under policy $\mu$. (Lillicrap et al., 2015; Satheeshbabu et al., 2020). Given the explanations provided, DRL holds significant potential for application in solving control engineering problems. Although DDPG provides a model-free solution, careful consideration is required when designing its hyperparameters and selecting suitable architectures for the actor and critic networks.

## 3 Proposed Learning-based Controller

This section starts with developing a customized Modified Transpose Jacobian algorithm for continuum robots. Subsequently, the process of designing an optimal adaptive gain-tuning system via DRL is presented.

### 3.1 Customized MTJ Control Algorithm for TDCRs

The MTJ control strategy aims to estimate the system dynamics using the previous time step control input in the task-space, achieving a performance similar to feedback linearization in model-based controllers within the model-free Transpose Jacobian (TJ) control algorithm (Moosavian and Papadopoulos, 2007; Craig, 2018). In this research, a customised MTJ control algorithm for TDCRs is proposed by incorporating the NSPO. The modification in the TJ structure involves adding a modification term, represented by the vector $\boldsymbol{h} = [h_x \quad h_y \quad h_z]^T$ to the TJ control input equation, If the position error vector in the task-space is defined as $\boldsymbol{e} = [e_x \quad e_y \quad e_z]^T$, the MTJ control input vector is expressed by the following equation:

$$\boldsymbol{T}_{\boldsymbol{MTJ}} = \boldsymbol{J}^T\left[\boldsymbol{K}_P\boldsymbol{e} + \boldsymbol{K}_I\int \boldsymbol{e}\,dt + \boldsymbol{K}_D\dot{\boldsymbol{e}} + \boldsymbol{h}\right] \tag{3}$$

The parameters in the above equation, which are the control gains ($\boldsymbol{K_i}$), are considered diagonal matrices and defined as follows:

$$\boldsymbol{K_i} = \begin{bmatrix} k_{i_x} & 0 & 0 \\ 0 & k_{i_y} & 0 \\ 0 & 0 & k_{i_z} \end{bmatrix} \quad , \; i = P\,, I\,, D \tag{4}$$

The modification term ($\boldsymbol{h}$) is calculated as follows:

$$\boldsymbol{h}_{(t)} = \boldsymbol{K}\boldsymbol{\mathcal{F}}_{(t-\Delta t)} \tag{5}$$

where $\boldsymbol{\mathcal{F}}_{(t-\Delta t)}$ is the previous time step control input in the task-space, and $\boldsymbol{K}$ is a diagonal matrix defined as:

$$\boldsymbol{K} = \begin{bmatrix} k_x & 0 & 0 \\ 0 & k_y & 0 \\ 0 & 0 & k_z \end{bmatrix} \tag{6}$$

The diagonal elements of matrix $\boldsymbol{K}$ are computed using the following equation:

$$k_i = exp\left[-\left(\frac{|e_i|}{e_{max_i}} + \frac{|\dot{e}_i|}{\dot{e}_{max_i}}\right)\right], i = x\,, y\,, z \tag{7}$$

where $e_{max_i}$ is the position error sensitivity threshold and $\dot{e}_{max_i}$ is the velocity error sensitivity threshold for activating the modification term. Ultimately, the MTJ algorithm preserves the advantages of the TJ strategy, such as structural simplicity, low computational cost, and a model-free nature, while addressing issues like noise sensitivity, amplification, and weaknesses in traversing fast paths. The described structure has been proven stable based on Lyapunov's stability theorems, ensuring asymptotic stability for the algorithm (Moosavian and Papadopoulos, 1997). If the vector of generalized forces in the joint-space (tendon tensions) for customized MTJ is defined as $\boldsymbol{T^+}$, the control input of the proposed strategy is expressed as:

$$\boldsymbol{T^+} = \boldsymbol{T_{MTJ}} + [\boldsymbol{I} - \boldsymbol{J}^T\boldsymbol{J}^{-T}]\,\boldsymbol{\zeta} \tag{8}$$

According to Equation 2, This formulation ensures that the tendons of the robot remain under tension (do not slack) while maintaining the control objective. The null-space adjustment vector ($\boldsymbol{\zeta}$) is determined by solving an optimization problem that ensures the cable tensions remain within their allowable range while minimizing additional forces. The optimization objective is typically to minimize $||\boldsymbol{\zeta}||^2$, which corresponds to minimizing unnecessary energy consumption. Constraints are applied to guarantee that the resulting cable tensions ($\boldsymbol{T^+}$) satisfy $T_{min} \leq \boldsymbol{T^+} \leq T_{max}$. Numerical methods, such as quadratic programming (e.g. fmincon in MATLAB), are used to efficiently compute $\boldsymbol{\zeta}$ while adhering to these constraints.

### 3.2 Optimal Adaptive Gain-Tuner System Design via Deep Reinforcement Learning

Given the explanations provided, this study uses DRL for online tuning of the gains of the model-free MTJ controller. Consequently, the DRL agent is responsible for determining the appropriate values for these gains in real-time. Figure 1 shows the block diagram related to the use of DRL in the MTJ control strategy. Here, the DRL agent's task is to determine the suitable values for the control gains in the $\boldsymbol{K}_P$, $\boldsymbol{K}_I$, and $\boldsymbol{K}_D$ matrices in real-time.

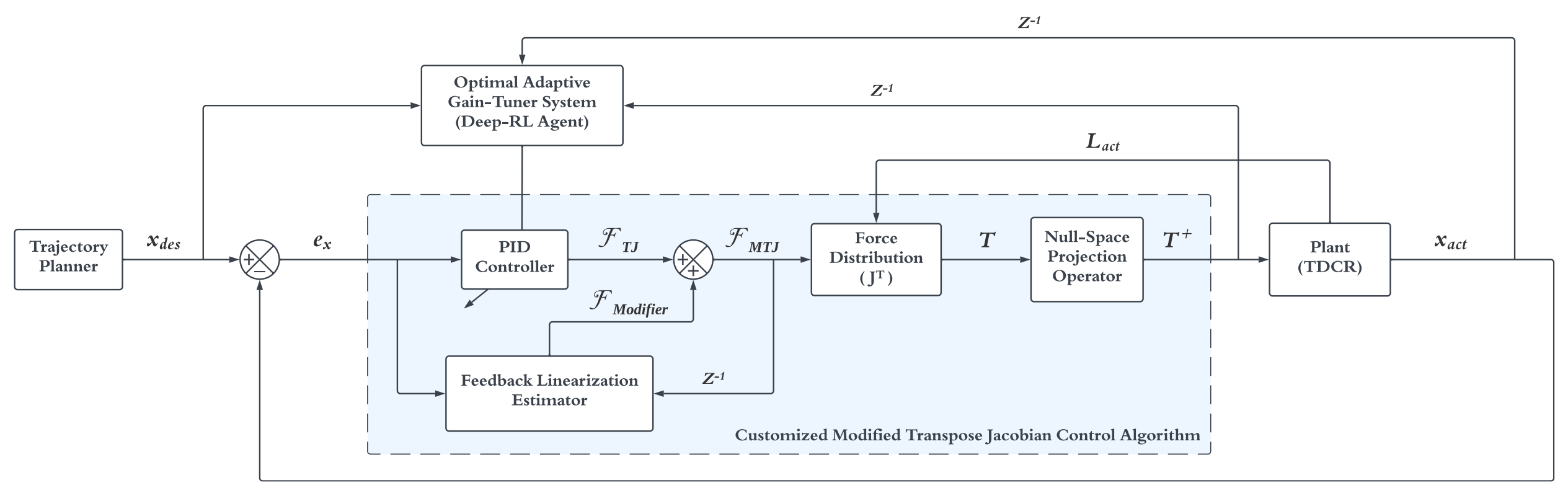

FIGURE 1
Block diagram of the proposed DRL-MTJ control strategy for TDCRs.

In the application of DRL to engineering problems, the reward function, state-space (observations), and action-space must first be defined. The following sections describe these elements for the problem of optimally tuning the MTJ controller gains for the position control of the end-effector in the TDCRs.

**1) State Space:** After trial and error and consideration of various variables, the position of the end-effector, position error, and the joint-space forces (tendon tensions) are defined as the system states:

$$States = \{x, y, z, e_x, e_y, e_z, T_1^+, T_2^+, T_3^+, T_4^+, T_5^+, T_6^+\} \tag{9}$$

**2) Action Space:** Given the objective for the DRL agent (to optimally tune the controller gains), the actions, or outputs of the DRL agent, are defined as:

$$Actions = \{K_{P_x}, K_{I_x}, K_{D_x}, K_{P_y}, K_{I_y}, K_{D_y}, K_{P_z}, K_{I_z}, K_{D_z}\} \qquad (10)$$

**3) Reward Function:** Based on the information obtainable from the environment (plant), the reward function is defined as a combination of the sum of squared errors (SSE) and a penalty function related to the controller gains:

$$Reward\ Function = -[SSE + 10\, f_G]$$

$$f_G = \sum_{i=x,y,z} [(K_{P_i} \leq K_{I_i}) + (K_{P_i} \leq K_{D_i}) + (K_{I_i} \leq K_{D_i})] \qquad (11)$$

In the above equation, $f_G$ is a function of the controller gains composed of Boolean variables, where each term can be either zero or one, representing whether the conditions on the gains are met.

**4) Actor and Critic Network Structures:** The study employed Multi-Layer Perceptron (MLP) neural network architectures for both the actor (policy approximator) and critic (value function approximator) components. The actor network, responsible for determining the optimal actions, processes the input state through several layers. Specifically, it starts with an input layer (observation), followed by three fully-connected layers each with 36 neurons and ReLU activation functions. The final layer is a fully-connected layer with 9 neurons, followed by a Tanh activation function and a scaling layer to produce the action output. Conversely, the critic network evaluates the value of state-action pairs. It takes the state and action as inputs, which are processed through a series of fully-connected layers with ReLU activations. The state input goes through two fully-connected layers each with 36 neurons and ReLU activations, while the action input is processed by one fully-connected layer with 36 neurons and a ReLU activation. These streams are then concatenated and passed through two more fully-connected layers, each with 36 neurons and ReLU activations, before reaching the output layer. The final output layer is a single neuron that represents the Q-value, indicating the value of the state-action pair.

To enhance the robustness of the neural network resulting from the execution of the DDPG algorithm, which maps the state-space to the action-space (actor), the reference point in the control loop and the initial conditions of the system (tendon tensions at the start of the simulation) are randomized in each episode.

# 4 Obtained Results

The obtained results from the learning process are presented in this section. The trained agent is ultimately employed as an optimal adaptive gain-tuning system in the simulation environment, and its performance in trajectory tracking is evaluated.

## 4.1 Learning Process Results

Figure 2-A shows the changes in discounted rewards per episode and their averages, as obtained by the DRL agent. Since all rewards in the defined reward function are negative, the ideal outcome would be to find a policy that results in a reward of zero throughout the episode. As depicted, after approximately 700 episodes, the DRL agent has nearly succeeded in estimating the optimal policy.

The graph in Figure 2-B presents the number of steps taken in each episode by the DRL agent and their averages. The results of this graph provide additional evidence for the success of the DRL agent in estimating the optimal policy. After about 700 episodes, the number of steps taken in each episode increases. This indicates that the episode termination condition (defined based on unfavorable conditions for the DRL agent) has not been activated.

Figure 2-C compares the average sum of discounted rewards with the value function estimated by the critic network. This graph provides key evidence of the DRL agent's success in estimating the optimal policy. According to the figure, the value function estimated by the critic network reaches a steady state after about 400 episodes, suggesting that the expected return of rewards received during each episode has stabilized. Referring to the graph of the average rewards, after about 700 episodes, the actor network has succeeded in finding the optimal policy, as the average sum of discounted rewards has nearly equaled the output of the critic network (which represents the value function or the expected return of rewards received during each episode). The hyperparameters considered for the algorithm are presented in the table of Figure 2-D.

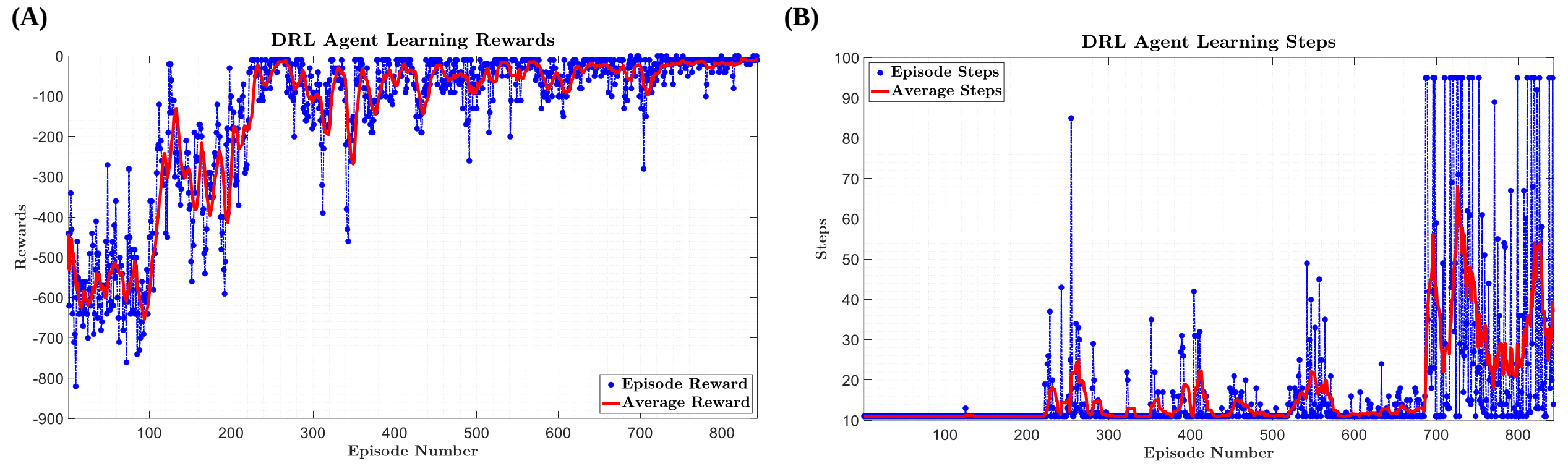

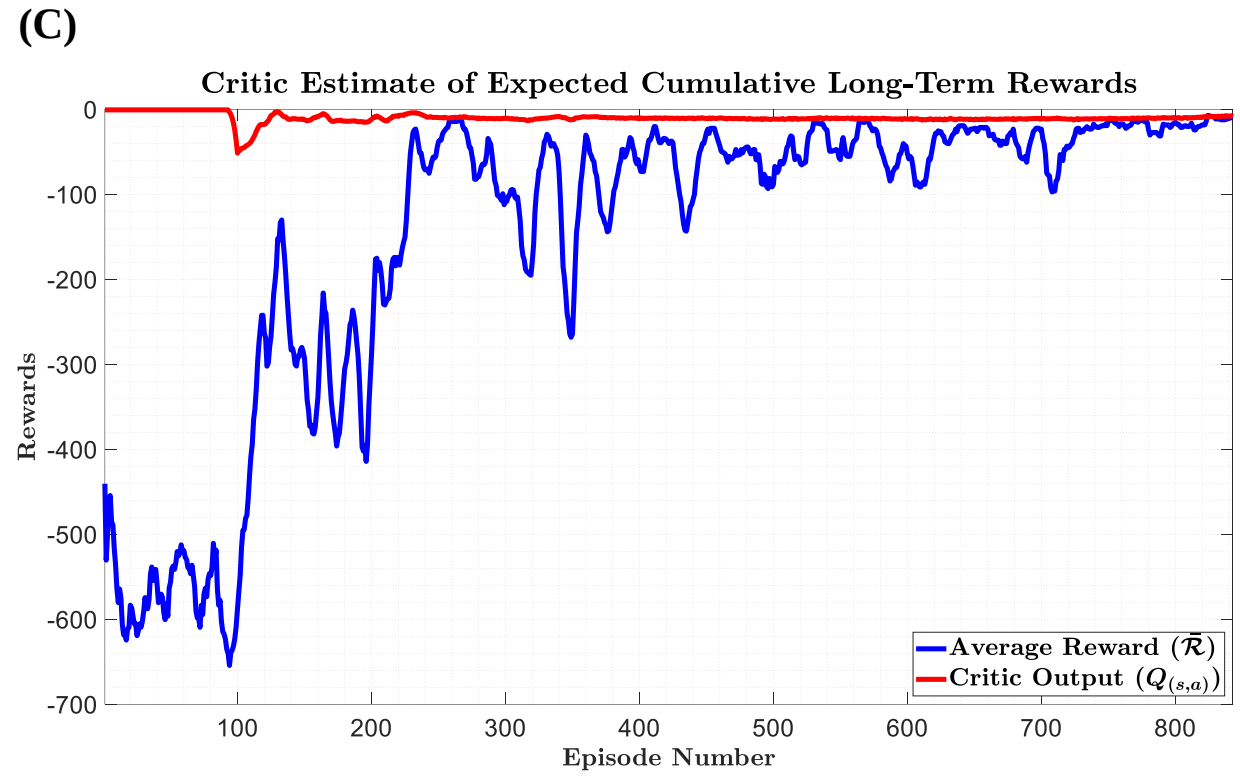

(D)

| Hyperparameter | Value |
|---|---|
| Experience Buffer Length | $10^6$ |
| Target Smooth Factor | $10^{-3}$ |
| Mini Batch Size | 1024 |
| Standard Deviation of Noise | 0.5 |
| Standard Deviation Decay Rate | $10^{-5}$ |
| Sample Time | 0.01 s |
| Episode Time | 3 s |
| Discount Factor | 0.99 |
| Maximum Number of Episodes | 1000 |

FIGURE 2
**(A)** Changes in discounted rewards received in each episode and their average. **(B)** Changes in the number of steps taken in each episode by the DRL agent and their average. **(C)** Comparison of the average sum of discounted rewards with the value function estimated by the critic network. **(D)** Values considered for the hyperparameters of the DDPG algorithm.

## 4.2 Simulation Results

To evaluate the performance quality of the DRL agent in tuning the controller gains, the trajectory designed by (Maghooli et al., 2023) is considered using the following equations:

$$\begin{cases} x_d = [0.2 + 0.025\cos(14t)]\sin(t) \\ y_d = [0.2 + 0.025\cos(14t)]\sin(t)\cos(t) \\ z_d = 0.2\, linsmf\left(\sqrt{0.4^2 - x^2 - y^2}, [0.25, 0.4]\right) + 0.2 \end{cases} \quad (12)$$

where $linsmf$ is a linear S-shaped fuzzy membership function. This allows for a fair comparison between the performance of the FIS and DRL in the position control problem of the TDCR. Figure 3-A illustrates the model used in the MATLAB environment and the trajectory followed by the continuum robotic arm. The main part of the plotting code is derived from (Rao et al., 2021). Figure 3-B shows the 3D path resulting from the considered trajectory.

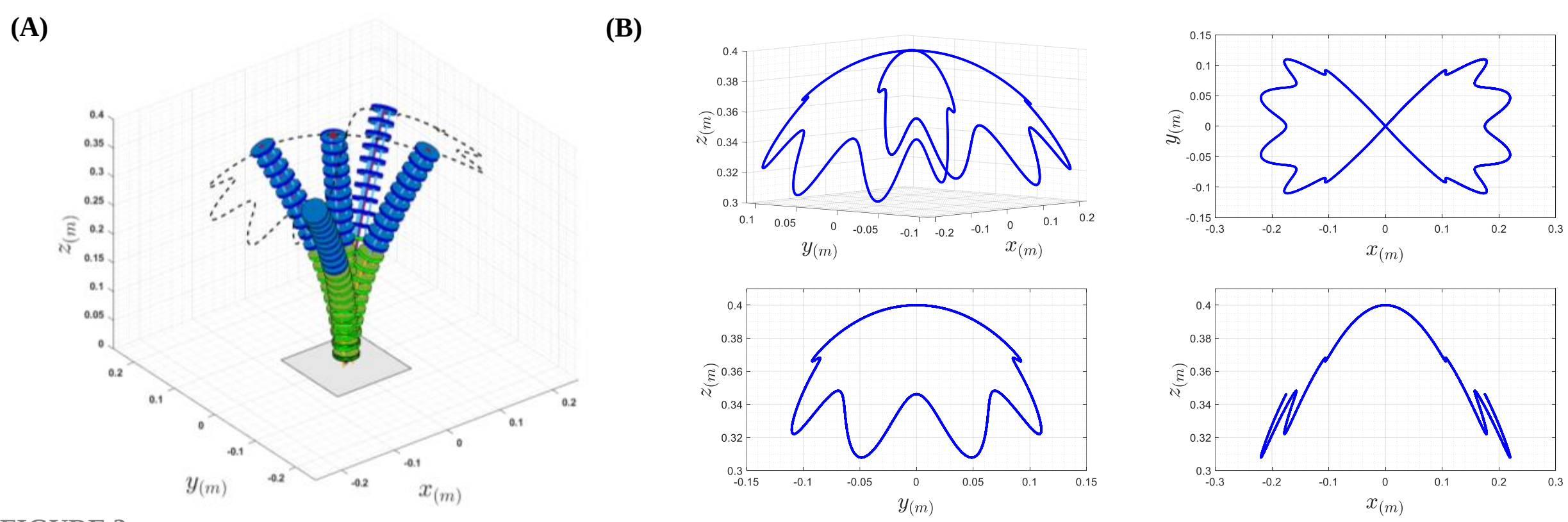

FIGURE 3
**(A)** A visualization of the robot movement in the MATLAB environment. **(B)** Reference path obtained from the designed trajectory for the TDCR in 3D Space.

By simulating the performance of the designed control systems to follow this path using the PCC model, the following results were obtained. In Figure 4, the trajectory-tracking quality of the MTJ controller with gains tuned by the DRL agent and the FIS is shown. Figure 4-A presents the graphs of the controller gains, adjusted in real-time by the DRL agent during trajectory tracking, compared to those provided by the FIS. The results for the x-direction are displayed, with similar results obtained for the other coordinates. A notable observation in the results is the variation in control gains by the DRL agent compared to the FIS over the simulation period. Specifically, the FIS provides nearly constant gains for a specific path throughout the duration, while the DRL agent updates the gains at each time step, striving to provide the most suitable gains for the current state of the robot. To better understand the performance of the controllers, the root mean square error (RMSE) is calculated throughout the simulation.

The tendon tension graphs for both strategies are shown in Figure 4-B. As observed, the tendon tensions are almost within the same range in terms of magnitude, indicating better management by the DRL agent compared to the FIS in optimally tuning the control gains and minimizing the error. According to the obtained values presented in Figure 4-C, using DRL compared to the FIS halves the RMSE in trajectory tracking, with values of 14.90 mm for FIS and 7.45 mm for DRL.

(A)

(B)

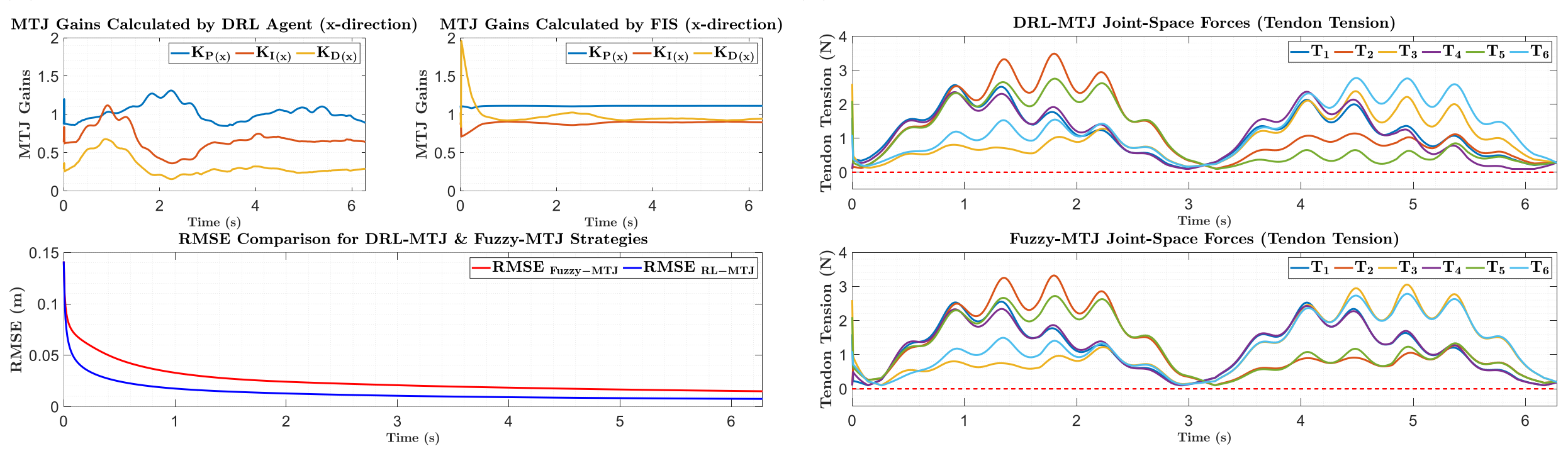

(C)

| RMSE | DRL-MTJ (mm) | Fuzzy-MTJ (mm) |
|---|---|---|
| x-direction | 5.77 | 6.66 |
| y-direction | 4.45 | 13.10 |
| z-direction | 1.55 | 2.55 |
| error norm | 7.45 | 14.90 |

FIGURE 4
**(A)** Comparison of controller gains and trajectory-tracking quality between Fuzzy-MTJ and DRL-MTJ strategies. **(B)** Comparison of tendon tension for both strategies. **(C)** Comparison of the RMSE for both strategies.

## 5 Experimental Implementation

This section begins by introducing the mechanical structure, electronic hardware, actuators, and sensors of the continuum robot developed in the ARAS laboratory. Subsequently, the challenges of transferring learning results from the simulation environment to the real world are discussed, and the validation of the Jacobian matrix for force distribution in the control loop is examined. Finally, the outcomes of transferring learning from the simulation to the real-world environment are evaluated.

### 5.1 Introduction to Continuum Robotic Arm Setup

The continuum robotic arm developed at the ARAS robotics laboratory (Robo-Arm) is a tendon-driven system with external actuation, as shown in Figure 5. The main components of the system are described below:

- **Backbone:** The backbone forms the main structure of the arm and is made of a nickel-titanium alloy (Nitinol). Nitinol is a shape-memory alloy, and its super-elasticity is the primary reason for its use as the central backbone of the system.
- **Robot Main Board:** The robot board serves as an interface between the computer and the system's actuators and sensors. All control commands to the servomotors and data received from the load cells are transmitted via the board through serial communication between the robot and the computer. The only exception is the cameras, whose data is directly transferred to the computer via USB ports.
- **Spacer Disks:** Spacer disks made of plexiglass are placed along the backbone to guide the tendons parallel to the central backbone. These disks also convert the tendon tension into a concentrated torque at the end of each segment, where the tendons attach to the backbone.
- **Tendons:** The tendons, with a maximum allowable tension of 394 Newtons, transfer force and ultimately convert it into concentrated torque at the end of each segment. When selecting the tendon material, inextensibility and flexibility are important characteristics, in addition to high maximum tension, as these significantly affect the system's power transmission performance.

- **Servomotors:** The actuators for the continuum robotic arm are Dynamixel servomotors (model AX-12A), which offer two modes: joint and wheel. These modes allow for position ($\theta$) and velocity ($\dot{\theta}$) control.
- **Load Cells:** Real-time information on tendon tension is essential for the kinetic control of the continuum robotic arm. By using load cells and implementing an inner loop to regulate tendon tension, the system can be kinetically controlled. The selected load cells have a maximum force capacity of 30 kg-force, which is approximately 294 Newtons.
- **Cameras:** To determine the real-time position of the continuum robotic arm's end-effector, two cameras are used to observe the robot's movement in the $xz$ and $yz$ planes. While real-time position information of the end-effector can be obtained through forward kinematics, factors such as increased computational load, potential delays in calculating these equations within the control loop, and structural and parametric uncertainties (e.g., friction, backlash, elasticity, hysteresis) can cause discrepancies between the calculated and actual positions of the end-effector. Therefore, the system uses two A4Tech cameras (30 fps), models PK-750MJ and PK-710MJ. Both models operate at 5 volts and 150 milliamps and can be easily connected to a computer via USB ports.

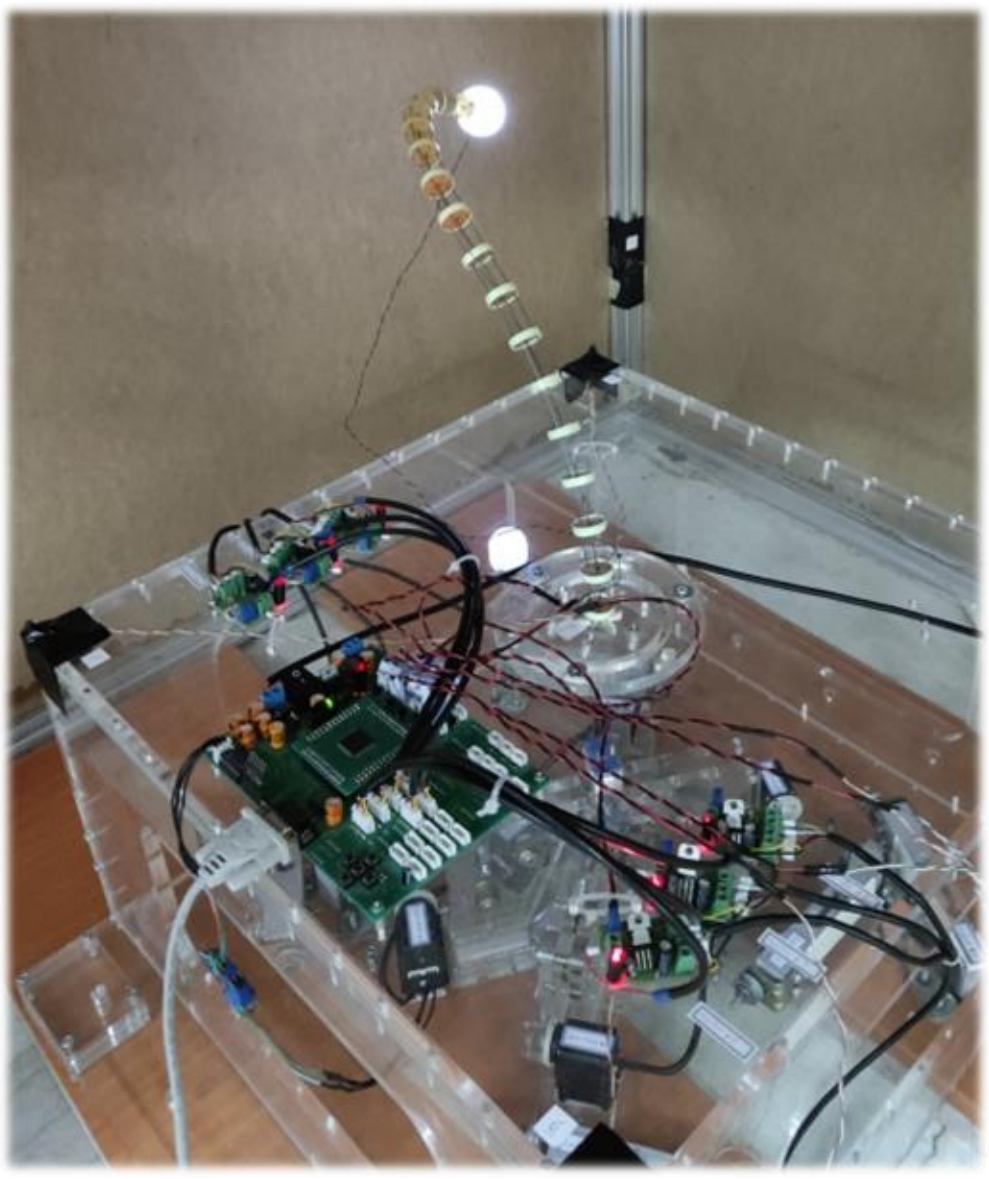

FIGURE 5
Image of the continuum robotic arm developed at the ARAS Robotics Lab.

### 5.2 Sim-to-Real Gap Considerations

One key difference between simulation and real-time implementation of this robotic system is the nature (dimension) of the input signals to the plant. In the simulation environment, the input to the system is the tendon tensions, and kinetic control is performed by determining these inputs. Conversely, in the hardware setup of the Robo-Arm, kinematic actuators (Dynamixel servomotors) are employed. If the input to the system is directly the motor position or velocity, a kinematic control strategy is implemented, which is not ideal for a continuum robot (Centurelli et al., 2022). This claim is supported by three reasons:

Firstly, kinematic control does not account for tendon tension. If the robot body or end-effector collides with the task-space or becomes stuck in the null-space, the controller cannot issue the correct commands to resolve these issues. Secondly, without information on tendon tension, the kinematic controller will not be aware if the tension increases beyond the tendon tolerance thresholds. This can lead to tendon rupture, damage to the spacer disks, or even damage to the robot backbone. Thirdly, the use of the Jacobian transpose as the force distributor is only possible with kinetic control. According to the equation $\boldsymbol{T} = \boldsymbol{J}^T \boldsymbol{\mathcal{F}}$, which maps forces from the task-space to the joint-space, $\boldsymbol{T}$ (input to the system) is in the form of forces. With kinematic control, using the Jacobian transpose for force distribution is not feasible. Instead, the inverse Jacobian ($\boldsymbol{J}^{-1}$) must be used according to the equation $\dot{\boldsymbol{L}} = \boldsymbol{J}^{-1}\dot{\boldsymbol{X}}$. Using $\boldsymbol{J}^{-1}$ in closed-loop control poses a significant risk and may cause the control algorithm to become unstable near singularity points (typically at the boundaries of the task-space).

Based on these reasons, it is evident that the appropriate strategy for position control of the TDCR is kinetic control. Implementing kinetic control, despite having kinematic actuators, involves using a cascaded control structure and creating an inner loop to adjust tendon tensions. In this structure, feedback from load cells is used to calculate the tendon tensions, which are then compared to the desired tension (output of the Jacobian transpose). The tension error is fed into the inner-loop controller (a PI controller), and finally, the command to adjust the motor angle is sent to the servomotor. The block diagram of the proposed strategy for kinetic control of the TDCR is shown in Figure 6.

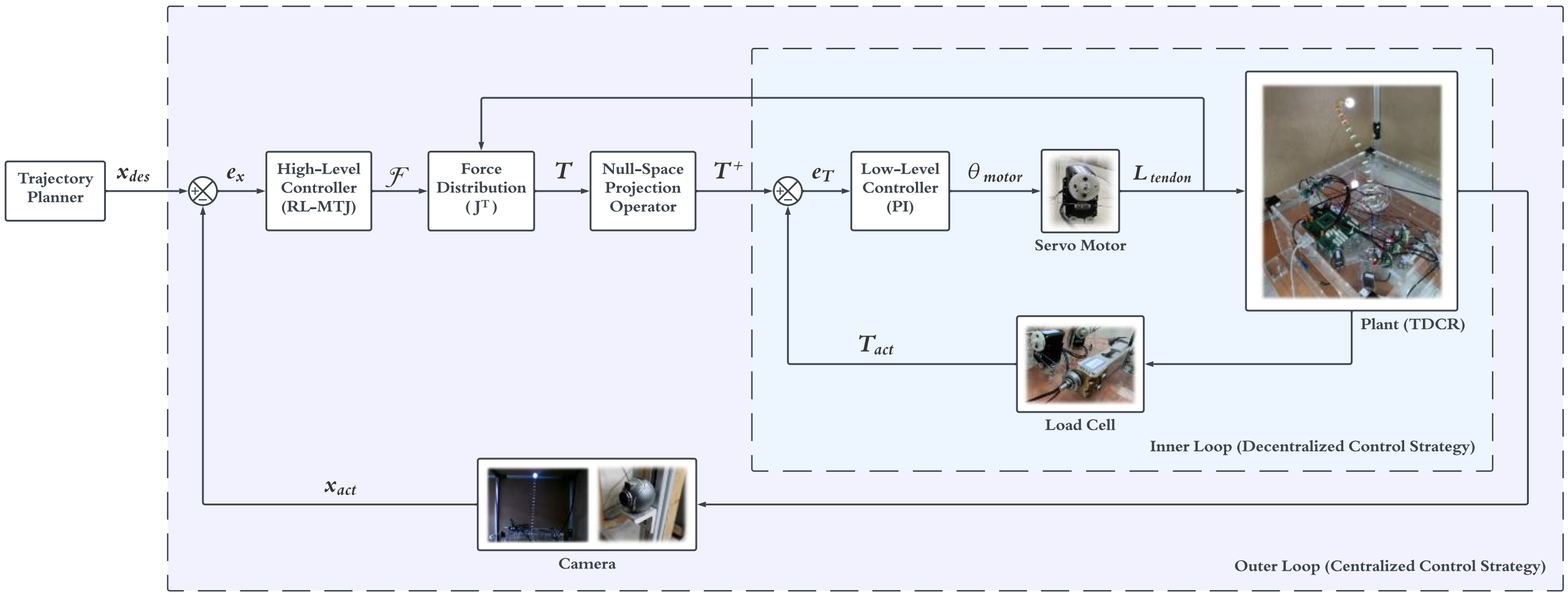

**FIGURE 6**
Proposed cascade control structure for kinetic control of the TDCRs in the task-space. Using a decentralized PI as the low-level controller (inner-loop) and a DRL-MTJ as the high-level controller (outer -loop).

### 5.3 Jacobian Validation for Force Distribution

As previously explained, having the Jacobian matrix for force mapping from the task-space to the joint-space is essential for control in the task-space. This research relies on using the closed form of the Jacobian matrix, which has shown satisfactory results in simulations. Before using these equations, it is necessary to ensure their accuracy and consistency with Robo-Arm for force distribution during real-time implementation.

Before addressing the Jacobian matrix, the validation of the forward kinematics equations (mapping tendon lengths to the end-effector position coordinates) is examined. Harmonic inputs with phase differences are applied to the servomotor angles, and the instantaneous position of the end-effector is recorded by cameras. The tendon lengths (calculated as the product of harmonic inputs and the servomotor pulley radius) are then input into the forward kinematics equations, and the output is compared with the camera measurements. Figure 7-A shows the comparison between the camera output and the forward kinematics output for each coordinate of the end-effector in the task-space. Based on the computed RMSEs, the forward kinematics equations, considering the structural and parametric uncertainties of the system, are accurate enough for calculating the Jacobian matrix for force distribution and implementation on hardware.

After validating the forward kinematics, the Jacobian matrix (which maps the rate of change in tendon lengths to the end-effector velocity) is examined. The derivative of the tendon lengths with respect to time is calculated and used as input for the Jacobian matrix, and the output is compared with the derivative of the camera measurements (representing the instantaneous velocity of the end-effector). Figure 7-B shows the comparison between the derivative of the camera output and the velocity obtained from the Jacobian for each coordinate of the end-effector in the task-space. According to the results, the Jacobian matrix is accurate enough for implementation on hardware, and its transpose can be used as the force distributor in the MTJ controller structure. The comparison of the RMSEs of the forward kinematics and Jacobian with camera data is presented in Figure 7-C.

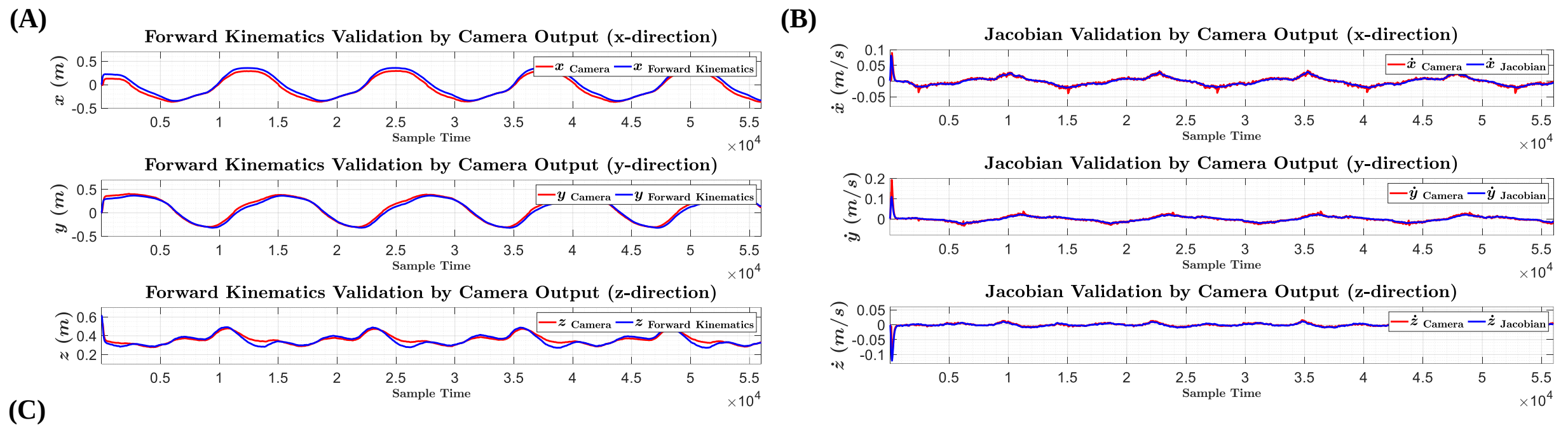

(C)

| RMSE | Forward Kinematics (mm) | Jacobian (mm/s) |
|---|---|---|
| x-direction | 59.01 | 3.92 |
| y-direction | 38.42 | 4.68 |
| z-direction | 24.21 | 2.98 |
| error norm | 74.46 | 6.79 |

**FIGURE 7**
**(A), (B)** Validation results of Forward Kinematics equations and Jacobian matrix using camera data. **(C)** Obtained RMSE from validation process.

## 5.4 DRL Agent Sim-to-Real Transfer Results

To successfully implement the MLP neural network obtained as the Actor on the continuum robotic arm hardware (Robo-Arm), derived from executing the DDPG algorithm in the simulation environment on a PCC model, the simulation model must closely approximate the physical system. This requirement is partially met by ensuring the accuracy of the mass and geometric parameters of the system in the model. However, due to the presence of uncertainties such as friction, hysteresis, and other factors that are challenging to model precisely, it is expected that the results of implementing the Actor network on the robot will differ somewhat from the simulation. The more effort that is put into accurately modeling these parameters, the smaller this discrepancy will be.

To evaluate and compare the performance of the DRL agent in tuning the controller gains, a circular trajectory in the horizontal plane is considered for a fair comparison between DRL and the FIS strategies. The trajectory is defined as follows:

$$\begin{cases} x_d = 0.15\sin(0.1t) \\ y_d = 0.15\cos(0.1t) \\ z_d = 0.48 \end{cases} \tag{13}$$

By implementing the designed control systems on the Robo-Arm, the following results are obtained. In. Figure 8, the trajectory-tracking quality of the MTJ controller with gains tuned by the DRL agent and the FIS is shown. Figure 8-A presents the controller gains graph (x-direction), adjusted in real-time by the DRL agent during trajectory tracking, compared to those provided by the FIS. Similar to the simulation results, the changes in control gains obtained by the DRL agent are more significant than those by the FIS throughout the implementation period. In other words, the DRL agent makes greater efforts to provide more suitable control gains at each time step according to the robot's state, resulting in a lower RMSE in trajectory tracking.

Figure 8-B shows the tendon tension graphs for both strategies. The results indicate that the tendon tensions are almost within the same range, demonstrating better management by the DRL agent compared to the FIS in tuning the control gains and minimizing the error. To better illustrate the performance of the controllers, the obtained RMSE for each coordinate for both strategies are presented in Figure 8-C. According to the obtained values, using DRL compared to the FIS significantly reduces the RMSE in trajectory-tracking (38.10 mm for DRL and 52.60 mm for FIS).

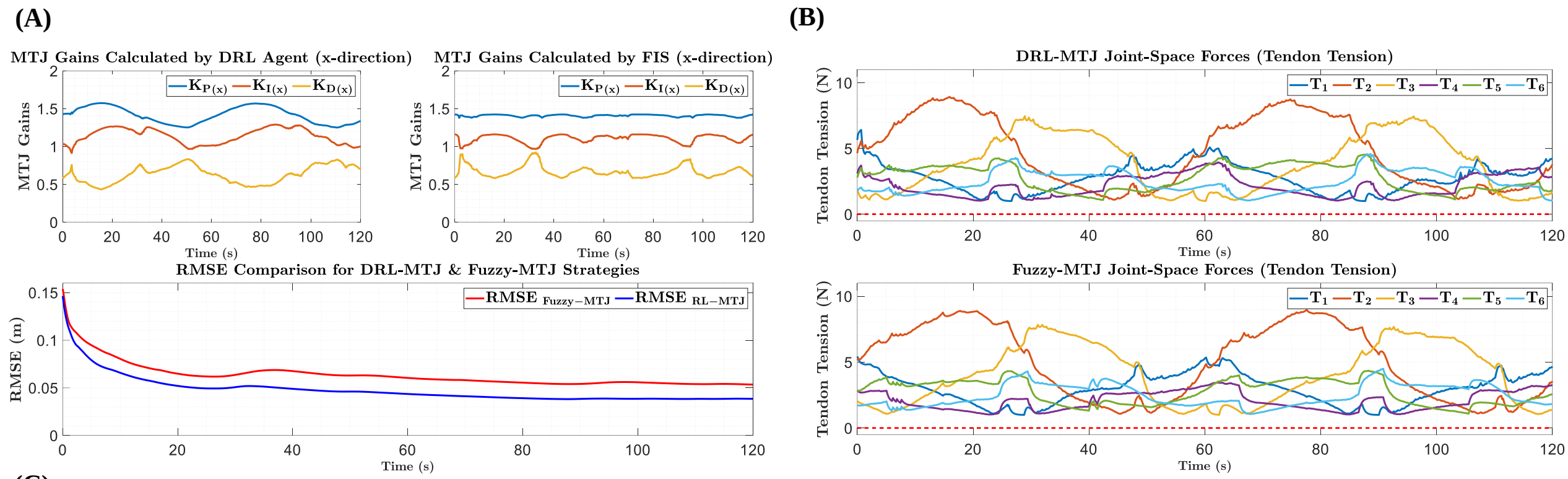

(C)

| RMSE | DRL-MTJ (mm) | Fuzzy-MTJ (mm) |
|---|---|---|
| x-direction | 26.22 | 40.71 |
| y-direction | 24.68 | 31.86 |
| z-direction | 12.15 | 9.69 |
| error norm | 38.10 | 52.60 |

**FIGURE 8**
**(A)** Comparison of controller gains and trajectory-tracking quality between Fuzzy-MTJ and DRL-MTJ strategies. **(B)** Comparison of tendon tension for both strategies. **(C)** Comparison of the RMSE for both strategies.

## 6 Discussion

In this study, the primary objective was to design an optimal adaptive gain-tuning system to enhance the performance of a customized MTJ controller for TDCRs. The results achieved in the trajectory-tracking problem, when compared to the application of a FIS for the same problem, demonstrate improvements in both simulation and real-world implementation. When comparing supervised learning methods (e.g., FIS) with semi-supervised methods (e.g., DRL), it can be stated that both approaches show satisfactory performance and require minimal prior knowledge of the system's behavior. Specifically, defining membership functions and the rule-base in a FIS necessitates knowledge of appropriate ranges for controller gains. On the other hand, defining states, actions, and rewards in DRL requires an understanding of how these variables affect system performance and their optimal selection within the problem's context. Notably, the FIS operates online from the outset and does not require prior training. However, the DRL agent can achieve appropriate online performance after a sufficient number of episodes and adequate training of the neural networks within its structure. Ultimately, based on the obtained results and the comparison of RMSE values, the DRL method demonstrates superior performance in tuning the proposed controller gains. Its application in the control of TDCRs results in more accurate following of the reference trajectory with reduced error. The simultaneous control of both position and orientation of a TDCR, leveraging the results of this paper to address shape constraints, has been independently explored in (Maghooli et al., 2024).

## 7 Conclusions

In this study, a learning-based control strategy was developed and validated by integrating a customized MTJ controller for TDCRs with the DDPG algorithm. The main contribution of this work lies in the effective Sim-to-Real transfer of control policies, enabling the model-free MTJ controller to achieve high-precision trajectory-tracking. The obtained results from both simulation and real-time implementation indicate that the optimal adaptive gain-tuning system significantly enhances controller performance, reducing the RMSE and improving the robustness of the control system. The success of this approach in both simulated and real-world environments underscores its potential for broader applications in medical devices, flexible manufacturing, and exploratory robotics. This work paves the way for more reliable and efficient deployment of TDCRs in real-world scenarios. Future work will focus on further optimizing the learning algorithms and exploring their application to shape estimation and control in continuum robotic arms.

## Supplementary Materials

A supplementary video demonstrating the robot calibration process, simulation results, and real-time implementation on the robot is available online at nima-maghooli.github.io.

## Data Availability Statement

The original contributions presented in this study are available upon request. Further inquiries can be directed to the corresponding author.

## Author Contributions

NM: conceptualization, investigation, methodology, software, validation, visualization, writing–original draft. OM: software, validation, writing–review and editing. MB: software, methodology, writing–review and editing. SAM: supervision, project administration, funding acquisition, resources, writing–review and editing.